\pdfoutput=1
\PassOptionsToPackage{table}{xcolor}
\documentclass[11pt]{article}
\usepackage[preprint]{acl}
\usepackage{times}
\usepackage{latexsym}
\usepackage[T1]{fontenc}
\usepackage[utf8]{inputenc}
\usepackage{microtype}
\usepackage{inconsolata}
\usepackage{graphicx}
\usepackage{amsmath}
\usepackage{amssymb}
\DeclareMathOperator*{\argmax}{arg\,max}
\usepackage{booktabs}
\usepackage{caption}
\usepackage{subcaption}
\usepackage{multirow}
\usepackage{pifont}
\usepackage{enumitem}
\usepackage{arydshln}
\usepackage{makecell}
\usepackage{placeins}
\usepackage{hyperref}
\hypersetup{
    colorlinks=false,
    hidelinks,
}
\title{LLMs Are Already Good Tutors: Training-Free Prompt Optimization for Pedagogical Math Tutoring}
\author{
\textbf{Unggi Lee}$^{1,\dagger}$ \quad \textbf{Minchul Shin}$^{2,\dagger}$ \quad \textbf{Yeil Jeong}$^{3,\dagger}$ \quad \textbf{Sookbun Lee}$^{4}$ \\
\textbf{Jeongsu Moon}$^{5}$ \quad \textbf{Kyungtae Joo}$^{5}$ \quad \textbf{Eunjoo Lee}$^{2}$ \quad \textbf{Hoilym Kwon}$^{6}$ \\[4pt]
$^{1}$Korea University Sejong Campus \quad $^{2}$Gyeonggi Institute of Education \quad $^{3}$Indiana University Bloomington \\
$^{4}$Opentutorials \quad $^{5}$Chosun University \quad $^{6}$Korea University Korean Studies Center \\[4pt]
$^{\dagger}$Corresponding authors \\
\texttt{codingchild@korea.ac.kr} \quad \texttt{minchul0582@gmail.com} \quad \texttt{yeilj@iu.edu}
}
\begin{document}
\maketitle
\begin{abstract}
Aligning LLMs for math tutoring typically requires RL-based training with multi-GPU infrastructure.
We investigate whether training-free prompt optimization-evolving only the system prompt via API calls-can serve as a practical alternative.
We adapt 7 published methods and propose 5 education-specialized methods, evaluating these 12 methods under 5 conditions on 2 OOD benchmark suites.
All 12 best-per-method configurations surpass the strongest RL-trained baseline ($R_\text{total}=0.633$), and our \textit{ParetoGrad} achieves the best Pareto balance across post-test solve rate, leak control, and helpfulness, rather than dominating any single component.
Behavioral analysis with an 82-code educational codebook reveals that training-free methods rely on teaching-knowledge patterns at 2-3$\times$ the rate of RL-trained models, with a compensating ${\sim}10$ percentage-point reduction in intent-level scaffolding.
We also find a task-dependent reasoning mode effect consistent across training-free and RL-based paradigms.
Our approach enables efficient development of pedagogically aligned LLM tutors with prompts alone and minimal compute.
\end{abstract}

\section{Introduction}

Large language models (LLMs) hold significant promise as personalized math tutors \citep{tack2022aiteacher,learnlmteam2025learnlm}, but aligning them with sound pedagogy-guiding students toward solutions without simply revealing answers-remains challenging \citep{chi2014icap}.
Recent RL-based alignment approaches \citep{dinucujianu2025pedagogicalrl,lee2026pedagogicalrl-thinking} train with multi-objective rewards (solve rate, leak prevention, helpfulness) using GRPO \citep{shao2024deepseekmath}, producing effective 7B tutors competitive with frontier models.
However, these approaches require multi-GPU infrastructure and thousands of training problems, creating a significant barrier for educators and researchers.

Meanwhile, automatic prompt optimization methods such as OPRO \citep{yang2024opro}, TextGrad \citep{yuksekgonul2024textgrad}, and GEPA \citep{agrawal2026gepa} have demonstrated that LLM behavior can be steered by evolving the system prompt without weight updates.
Yet these methods have never been applied to multi-turn educational dialog, where optimization must jointly balance scaffolding quality, answer leak prevention, and student learning outcomes.

\begin{figure}[t]
\centering
\includegraphics[width=\columnwidth]{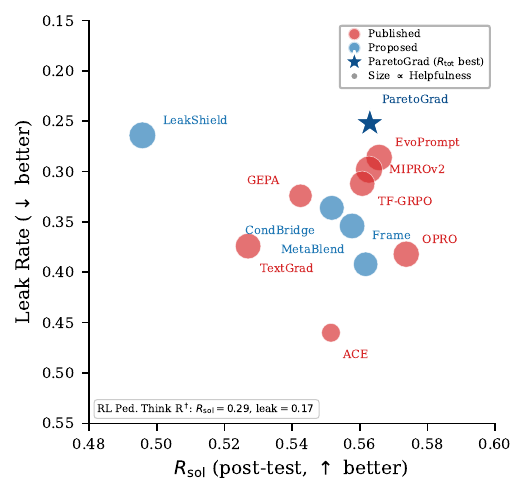}
\caption{Post-test solve rate ($R_\text{sol}$, K=8) vs.\ leak rate trade-off across the 12 listed training-free methods; bubble size reflects helpfulness. Our proposed \textit{ParetoGrad} ({\color[HTML]{0d4f8b}$\bigstar$}) achieves the best Pareto balance across the three objectives without updating model weights.}
\label{fig:tradeoff}
\end{figure}

\begin{figure*}[t]
\centering
\includegraphics[width=\textwidth]{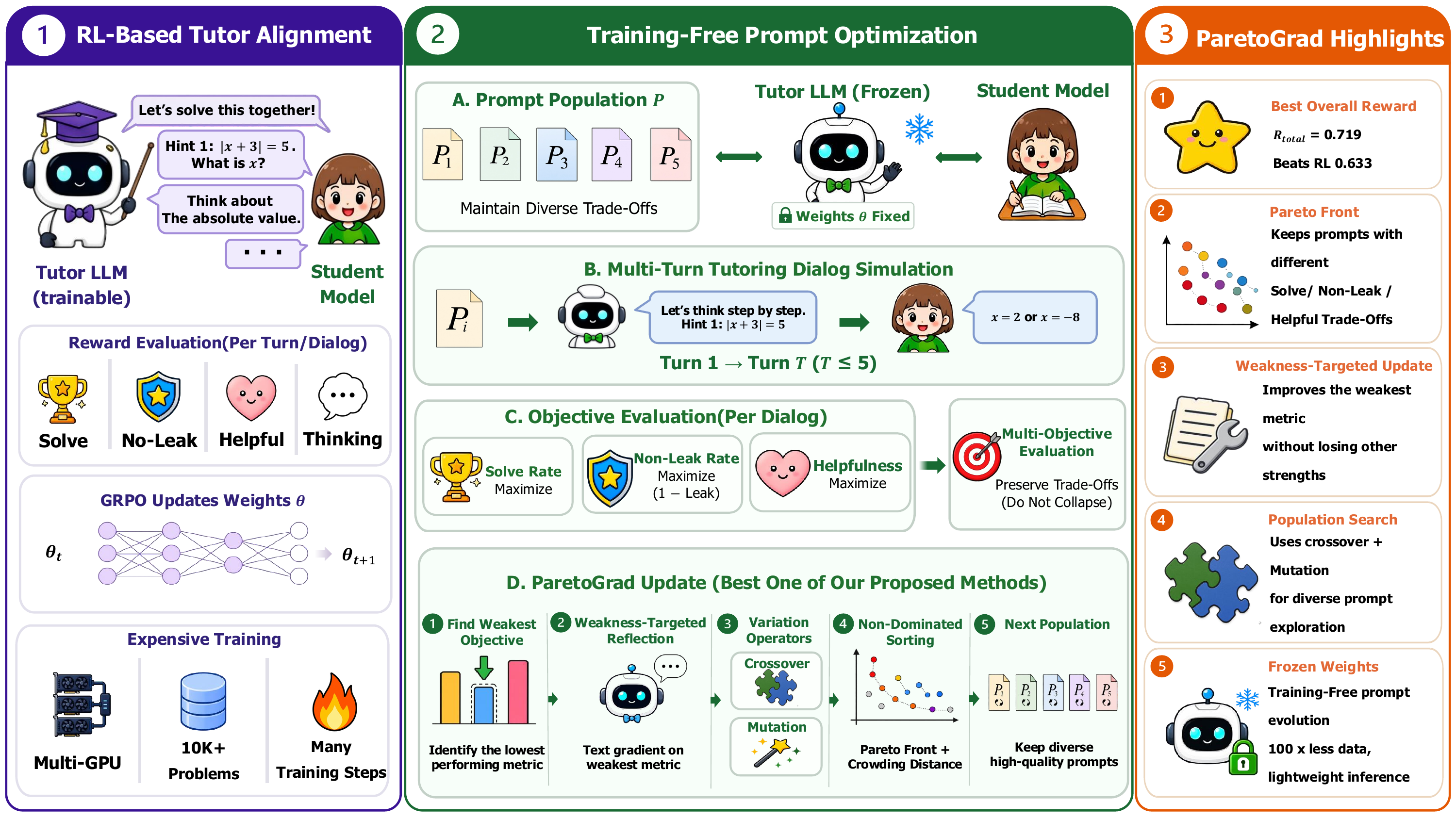}
\caption{\textit{Left} shows RL-based tutor alignment and \textit{center} shows our training-free prompt optimization, illustrated with \textit{ParetoGrad} as a representative instance; \textit{right} lists method highlights. RL updates tutor weights $\theta$ via GRPO over reward components (solve, non-leak, helpfulness, thinking), requiring multi-GPU training over 10K+ problems. In contrast, our approach keeps the tutor and student models frozen and evolves a population of system prompts $\mathcal{P}$ through four steps - (A) maintaining diverse prompt trade-offs, (B) simulating multi-turn tutoring dialog, (C) evaluating each prompt on three objectives ($R_\text{sol}$, $R_\text{leak}$, $R_\text{help}$), and (D) updating the population via weakness-targeted reflection, crossover and mutation, and non-dominated Pareto sorting with crowding distance. \textit{ParetoGrad} achieves the best Pareto balance across solve rate, leak control, and helpfulness with 100$\times$ less data and single-GPU inference, and all 12 listed training-free methods surpass the strongest RL-trained baseline.}
\label{fig:overview}
\end{figure*}

We investigate whether \emph{training-free} prompt optimization can serve as a practical alternative to RL-based tutor alignment (Figures~\ref{fig:tradeoff} and~\ref{fig:overview}).
We first adapt 7 published prompt optimization methods to multi-turn educational dialog.
Through extensive experimentation with diverse optimization strategies, we identify and propose 5 education-specialized methods that embed pedagogical priors - scaffolding, leak prevention, meta-optimization - directly into the prompt evolution process.
Using the same reward framework as PedagogicalRL-Thinking \citep{lee2026pedagogicalrl-thinking}, we evaluate all 12 methods under 5 conditions on 2 OOD benchmark suites \citep{lee2026openlearnlm,macina2025mathtutorbench}, yielding 792 evaluation runs.

Our key contributions are:
\begin{itemize}[nosep,leftmargin=*]
  \item We show that training-free prompt optimization can match or exceed RL-trained baselines on the same multi-objective reward using only inference-time compute.\footnote{\url{https://anonymous.4open.science/r/tf-openlearnlm-anon-C780/README.md}}
  \item We propose five education-specialized prompt optimization methods (\textit{ParetoGrad}, \textit{CondBridge}, \textit{LeakShield}, \textit{Frame}, \textit{MetaBlend}) inspired by empirical failure analysis of existing approaches.
  \item We provide the first adaptation of published prompt optimization methods to multi-turn educational dialog, together with a systematic behavioral comparison against RL-trained approaches using the same educational codebook.
\end{itemize}

\section{Method}

\subsection{Problem Formulation}

Given a frozen LLM $\mathcal{M}$ with fixed weights $\theta$, we seek a system prompt $P^*$ that maximizes tutoring quality:
\begin{equation}
    P^* = \argmax_{P} \; \mathbb{E}_{x \sim \mathcal{X}} \left[ R_{\text{total}}\!\left(\text{Dialog}(\mathcal{M}, P, x)\right) \right]
\end{equation}
Unlike RL-based approaches \citep{dinucujianu2025pedagogicalrl,lee2026pedagogicalrl-thinking} that update $\theta$ via GRPO, we keep $\theta$ frozen and evolve only $P$, requiring only black-box API access.
Each candidate prompt $P$ is evaluated by simulating multi-turn tutoring dialogs between the tutor $\mathcal{M}$ and a student model $\mathcal{S}$ for up to $T\!=\!5$ turns on problem $x$.

\subsection{Reward Design}

We adopt the reward framework from \citet{lee2026pedagogicalrl-thinking}.
Given a dialog $D = \text{Dialog}(\mathcal{M}, P, x)$, we compute four reward components:
\textbf{(1)} $R_{\text{sol}}(D)$: post-dialog student solve rate over $K\!=\!8$ attempts;
\textbf{(2)} $R_{\text{leak}}(D) \in \{0,1\}$: 1 if the tutor avoided leaking the answer, 0 otherwise (we report the leak rate $1 - R_{\text{leak}}$ in tables for interpretability; $\rho = -0.94$ between leak rate and $R_{\text{total}}$ across 137 method$\times$condition runs);
\textbf{(3)} $R_{\text{help}}(D) \in [0,1]$: pedagogical helpfulness;
\textbf{(4)} $R_{\text{think}}(D) \in [0,1]$: thinking quality (reasoning models only).
The total reward is:
\begin{equation}
\small
R_{\text{total}}(D) \!=\! \begin{cases}
\frac{R_{\text{sol}} + R_{\text{leak}} + R_{\text{help}}}{3} & \text{(no think)} \\[4pt]
\frac{R_{\text{sol}} + R_{\text{leak}} + R_{\text{help}} + R_{\text{think}}}{4} & \text{(think)}
\end{cases}
\end{equation}

\subsection{Adapted Published Methods}

We adapted 7 published prompt optimization methods to the tutoring domain, replacing their evaluation functions with our dialog simulation and reward $R_\text{total}$.
GEPA \citep{agrawal2026gepa} uses reflective prompt evolution with a Pareto frontier over $R_\text{sol}$, $R_\text{leak}$, and $R_\text{help}$.
ACE \citep{zhang2026ace} extracts reusable skills, which we redefine as pedagogical strategies such as scaffolding and questioning.
OPRO \citep{yang2024opro} uses an LLM-as-optimizer approach, where we include tutoring failure cases in the meta-prompt.
EvoPrompt \citep{guo2024evoprompt} applies genetic algorithms over a population of tutor system prompts $\{P_1, \ldots, P_N\}$.
TextGrad \citep{yuksekgonul2024textgrad} computes text-based gradients $\nabla_\text{text}$ from dialog quality feedback and updates $P_{t+1} = P_t - \nabla_\text{text}$.
MIPROv2 \citep{opsahlong2024miprov2} performs Bayesian selection over pedagogical instruction variants.
TF-GRPO applies the group relative scoring from GRPO \citep{shao2024deepseekmath} to rank prompt candidates by $R_\text{total}$ without updating $\theta$.

\subsection{Proposed Education-Specialized Methods}
\label{sec:proposed}

Error analysis of 19{,}500 dialogs from the adapted methods revealed three tutoring-specific failure patterns: \emph{answer leakage} (82.8\% of failures), \emph{rigid instruction following}, and \emph{lack of pedagogical structure} despite high $R_\text{total}$ \citep{chi2014icap}.
Inspired by these empirical observations, we designed 5 education-specialized methods targeting these failure patterns.
We present these methods below; additional exploratory variants are documented in Appendix~\ref{app:methods}.

\subsubsection{Pedagogical Scaffolding}
\textit{Frame} extends TextGrad \citep{yuksekgonul2024textgrad} with a post-gradient reframing pass:
\begin{equation}
P_{t+1} = \text{Refr}\!\left(P_t - \nabla_\text{text}(P_t, D)\right)
\end{equation}
where $\text{Refr}(\cdot)$ applies three linguistic transformations: (1) converting prohibitive language into behavioral guidance; (2) replacing brevity directives with substance-focused instructions; (3) injecting concrete tutoring exemplars.
In pilot analysis, reframing alone reduced leak rate from 0.891 to 0.761.

\subsubsection{Leak Prevention}
\textit{CondBridge} evaluates each candidate $P$ under both NoThink ($\mathcal{M}_\text{NT}$) and Think ($\mathcal{M}_\text{TH}$) conditions:
\begin{equation}
\small
P^* \!=\! \argmax_P \alpha R_\text{total}^{\text{NT}}\!(P) + (1\!-\!\alpha) R_\text{total}^{\text{TH}}\!(P)
\end{equation}
where $\alpha = 0.5$.
This dual-condition objective drives \textit{CondBridge} to the lowest leak rate (0.204) among all training-free methods at NoThink, while at Think Reward it reaches $R_\text{total}=0.691$.

\subsubsection{Dual-Objective}
\textit{LeakShield} uses two-stage optimization: Stage~1 (first 40\%) minimizes leakage; Stage~2 maximizes $R_\text{sol}$ and $R_\text{help}$ under a leak ceiling $\tau$:
\begin{equation}
\small
P^* \!=\! \argmax_P R_\text{sol}(P) \!+\! R_\text{help}(P) \;\text{s.t.}\; 1\!-\!R_\text{leak}(P) \!\leq\! \tau
\end{equation}
If leakage exceeds $\tau$ during Stage~2, the optimizer reverts to anti-leak updates.

\subsubsection{Meta-Optimization}
\textit{MetaBlend} collects the best prompts $\{P^*_1, \ldots, P^*_m\}$ from $m$ completed runs, extracts common structural patterns, and synthesizes an initial prompt $P_0$.
Phase~2 refines $P_0$ via TextGrad for 300 iterations, preserving patterns as constraints.

\subsubsection{Population-Gradient Hybrid}
\textit{ParetoGrad} maintains a population $\{P_1, \ldots, P_N\}$ with $N\!=\!5$ and applies NSGA-II non-dominated sorting across $(R_\text{sol}, R_\text{leak}, R_\text{help})$, yielding the best Pareto balance across the three objectives ($R_\text{total} = 0.719$ at Think NoReward).
Each generation produces offspring via weakness-targeted TextGrad, crossover, and mutation over 100 generations.

\section{Experiments}

\begin{table*}[!t]
\centering
\scriptsize
\setlength{\tabcolsep}{2pt}
\renewcommand{\arraystretch}{1.0}
\caption{In-domain BigMath and OOD benchmark results. Each training-free row shows its best of 5 conditions, encoded by the \textit{Th./Th.R/Prompt} indicators. $R_\text{sol}$ is the K=8 post-dialog solve rate; $R_\text{total}=(R_\text{sol}+R_\text{leak}+R_\text{help})/3$. \textit{ParetoGrad} achieves the best Pareto balance across the three components, and all 12 listed training-free methods surpass the strongest RL-trained baseline.}
\label{tab:main}
\begin{tabular}{@{}lcclcccccccccccc@{}}
\toprule
& & & & \multicolumn{4}{c}{\textbf{BigMath (ID)}} & \multicolumn{5}{c}{\textbf{OpenLearnLM (OOD)}} & \multicolumn{3}{c}{\textbf{MTBench}} \\
\cmidrule(lr){5-8} \cmidrule(lr){9-13} \cmidrule(lr){14-16}
\textbf{Method} & \textbf{Th.} & \textbf{Th.R} & \textbf{Prompt}
  & $\boldsymbol{R_\text{sol}}{\uparrow}$ & \textbf{Leak}$\downarrow$ & \textbf{Help}$\uparrow$ & $\boldsymbol{R_\text{total}}{\uparrow}$
  & \textbf{CK}$\uparrow$ & \textbf{PK}$\uparrow$ & \textbf{SK}$\uparrow$ & \textbf{Att}$\uparrow$ & \textbf{Avg}$\uparrow$
  & \textbf{Sc}$\uparrow$ & \textbf{Pd}$\uparrow$ & \textbf{Avg}$\uparrow$ \\
\midrule
\rowcolor{gray!20}
\multicolumn{16}{@{}l}{\textbf{Frontier Models (Zero-shot)}\textsuperscript{$\dagger$}} \\
GPT-5.2 (Ped.)         & \ding{51} & - & Ped. & .340 & \textbf{.000} & .440 & .593 & \textbf{8.08} & 8.46 & 8.63 & 8.68 & \textbf{8.46} & - & - & - \\
Claude-4-Opus (Ped.)    & \ding{51} & - & Ped. & .350 & .090 & .760 & .673 & 6.63 & \textbf{8.61} & \textbf{8.82} & 8.45 & 8.13 & - & - & - \\
DeepSeek-V3.2 (Ped.)   & \ding{51} & - & Ped. & .390 & .110 & .820 & .700 & 7.46 & 7.32 & 8.63 & \textbf{8.77} & 8.05 & - & - & - \\
\midrule
\rowcolor{gray!20}
\multicolumn{16}{@{}l}{\textbf{RL-Trained Models}\textsuperscript{$\dagger$}} \\
NoThink (RL)            & \ding{55} & \ding{55} & Gen. & .120 & .300 & .180 & .333 & 7.95 & 7.43 & 7.77 & 7.57 & 7.68 & - & - & - \\
Think NR (RL)           & \ding{51} & \ding{55} & Gen. & .281 & .180 & .730 & .604 & 7.95 & 7.43 & 7.76 & 7.79 & 7.73 & - & - & - \\
Think R (RL)            & \ding{51} & \ding{51} & Gen. & .284 & .182 & .764 & .621 & 7.95 & 7.43 & 7.76 & 7.71 & 7.71 & - & - & - \\
Ped.\ Think NR (RL)     & \ding{51} & \ding{55} & Ped. & .275 & .214 & .766 & .607 & 7.95 & 7.43 & 7.76 & 7.79 & 7.73 & - & - & - \\
Ped.\ Think R (RL)      & \ding{51} & \ding{51} & Ped. & .294 & \underline{.172} & .776 & .633 & \underline{7.99} & 7.43 & 7.77 & 7.86 & 7.76 & - & - & - \\
\midrule
\rowcolor{gray!20}
\multicolumn{16}{@{}l}{\textbf{Baseline}} \\
No optimization         & \ding{55} & \ding{55} & Gen. & .120 & .300 & .180 & .333 & 7.56 & 6.87 & 8.53 & 7.93 & 7.72 & 7.17 & 5.25 & 6.21 \\
\midrule
\rowcolor{gray!20}
\multicolumn{16}{@{}l}{\textbf{Training-Free: Published Methods (Adapted)}} \\
EvoPrompt    & \ding{51} & \ding{51} & Gen. & \underline{.566} & .286 & .847 & \underline{.711} & 7.62 & \underline{7.54} & \underline{8.69} & 8.50 & \underline{8.09} & 8.16 & \textbf{8.09} & 8.13 \\
MIPROv2      & \ding{51} & \ding{51} & Gen. & .563 & .298 & \textbf{.848} & .707 & 6.97 & 7.42 & 8.56 & 8.21 & 7.79 & \textbf{8.29} & 8.03 & \textbf{8.16} \\
TF-GRPO      & \ding{51} & \ding{51} & Gen. & .561 & .312 & .845 & .700 & 6.78 & 6.71 & 7.95 & 8.50 & 7.49 & 7.95 & 7.70 & 7.82 \\
GEPA         & \ding{51} & \ding{55} & Gen. & .542 & .324 & .841 & .686 & 6.41 & 6.87 & 7.27 & 8.43 & 7.24 & \underline{8.23} & 8.00 & 8.12 \\
OPRO         & \ding{51} & \ding{51} & Ped. & \textbf{.574} & .382 & .846 & .684 & 6.64 & 6.67 & 7.70 & 8.21 & 7.31 & 7.84 & 7.65 & 7.75 \\
TextGrad     & \ding{51} & \ding{55} & Ped. & .527 & .374 & .845 & .666 & 6.97 & 6.67 & 7.45 & 8.36 & 7.36 & 8.01 & 7.72 & 7.86 \\
ACE          & \ding{51} & \ding{55} & Ped. & .551 & .460 & .833 & .642 & 7.11 & 6.63 & 8.62 & \underline{8.57} & 7.73 & 8.21 & \underline{8.06} & \underline{8.13} \\
\midrule
\rowcolor{cyan!15}
\multicolumn{16}{@{}l}{\textbf{Training-Free: Proposed Methods (Ours)}} \\
ParetoGrad   & \ding{51} & \ding{55} & Gen. & .563 & .252 & .845 & \textbf{.719} & 6.92 & 6.55 & 7.99 & 8.07 & 7.38 & 7.77 & 7.79 & 7.78 \\
LeakShield   & \ding{51} & \ding{55} & Gen. & .496 & .264 & .847 & .693 & 6.97 & 6.87 & 7.12 & 8.00 & 7.24 & 7.37 & 7.87 & 7.62 \\
CondBridge   & \ding{51} & \ding{51} & Gen. & .552 & .336 & .843 & .691 & 6.88 & 6.71 & 7.91 & 8.14 & 7.41 & 7.82 & 7.84 & 7.83 \\
Frame        & \ding{51} & \ding{55} & Ped. & .558 & .354 & .845 & .683 & 6.97 & 6.83 & 7.62 & 7.93 & 7.34 & 7.89 & 7.72 & 7.81 \\
MetaBlend    & \ding{51} & \ding{51} & Gen. & .562 & .392 & .843 & .676 & 6.88 & 6.79 & 7.68 & 8.29 & 7.41 & 7.88 & 7.63 & 7.76 \\
\bottomrule
\end{tabular}
\par\smallskip
{\raggedright\scriptsize\textsuperscript{$\dagger$}Frontier/RL/Baseline rows show $\Delta$Solve in the $R_\text{sol}$ column as reported by \citet{lee2026pedagogicalrl-thinking,lee2026openlearnlm}; training-free rows use the K=8 post-test $R_\text{sol}$ defined above. \textbf{Bold} = best per column (for $R_\text{sol}$ and $R_\text{total}$, restricted to the 12 listed training-free methods due to the $R_\text{sol}$ metric difference). \underline{Underline} = best per column among non-Frontier rows (RL + 12 training-free) when distinct from the bolded entry, else 2nd best in that pool. Full appendix results in \S\ref{app:full-results}.\par}
\end{table*}

\subsection{Setup}

We match the setup of \citet{dinucujianu2025pedagogicalrl,lee2026pedagogicalrl-thinking}.
The tutor model is Qwen2.5-7B-Instruct for the NoThink condition with a maximum of 256 output tokens, and Qwen3-8B for all thinking conditions with 384 output tokens and a thinking budget of 1{,}024 tokens.
The student model is LLaMA-3.1-8B-Instruct with a maximum of 512 tokens.
Reward judgments are made by GPT-4o-mini \citep{zheng2023llmjudge}, and prompt improvements are proposed by GPT-4o as the reflection model.
Each dialog runs up to 5 turns under 5 conditions identical to \citet{lee2026pedagogicalrl-thinking}: NoThink (Qwen2.5-7B), Think NoReward and Think Reward (Qwen3-8B, with/without $R_\text{think}$), and their pedagogical-seed variants.
Whether each condition enables thinking, applies $R_\text{think}$, or seeds with a pedagogical prompt is encoded in Table~\ref{tab:main} as the \textit{Th./Th.R/Prompt} indicator triplet.

Optimization uses 100 BigMath \citep{albalak2025bigmath} problems filtered to medium-to-hard difficulty (student solve rate 1-60\%).
We evaluate on 2 OOD suites: OpenLearnLM \citep{lee2026openlearnlm} (4 sub-benchmarks) and MathTutorBench \citep{macina2025mathtutorbench} (2 sub-benchmarks), totaling 1{,}334 items.
Our approach requires only a single RTX 3090 for vLLM inference, compared to 4$\times$H100 GPUs for RL training \citep{lee2026pedagogicalrl-thinking}, with a small amount of API costs for GPT-4o-mini and LLaMA-3.1 via OpenRouter.
We conduct 792 evaluation runs in total.
We focus on the 5 proposed methods listed in Table~\ref{tab:main}; additional exploratory variants are in Appendix~\ref{app:full-results}.

\subsection{Main Results}
\label{sec:main-results}

Table~\ref{tab:main} presents performance on both in-domain and OOD benchmarks; for compactness each training-free method is shown with its best of 5 conditions ranked by $R_\text{total}$, and the corresponding configuration is encoded by the indicator triplet.

\subsubsection*{RQ1: Is training-free prompt optimization effective for math tutoring?}

Yes - and more strongly than initial estimates suggested.
All training-free methods improve over the unoptimized baseline ($R_\text{total}\!\approx\!0.33$); all 12 listed methods even surpass the strongest RL-trained model (Ped.\ Think R, $R_\text{total}=0.633$).
Our proposed \textit{ParetoGrad} (Think, no reward) reaches $R_\text{total}=0.719$, the most balanced result across the three objectives.
ParetoGrad does not dominate any single component but is uniformly strong, with $R_\text{sol}=0.563$ (3rd, tied with MIPROv2), Leak$=0.252$ (best), and Help$=0.845$ (top tier).
Specialists exist for individual components - \textit{OPRO} leads $R_\text{sol}$ (0.574) and \textit{CondBridge} attains the lowest leak (0.204) at NoThink - but neither balances all three components as well as ParetoGrad, illustrating the persistent solve-leak tension (Figure~\ref{fig:tradeoff}).

\subsubsection*{RQ2: Do education-specialized methods outperform general-purpose ones?}

The answer is nuanced under the completed evaluation.
On in-domain BigMath $R_\text{total}$, the top-12 split into 7 published-adapted and 5 proposed methods; the most balanced is our proposed \textit{ParetoGrad} (0.719), with published \textit{EvoPrompt} (0.711) and \textit{MIPROv2} (0.707) close behind on aggregate reward.
On OOD, published methods now lead OpenLearnLM-Avg (\textit{EvoPrompt} 8.09) and MathTutorBench-Avg (\textit{MIPROv2} 8.16), with proposed methods narrowly behind (\textit{CondBridge} OL-Avg 7.41; \textit{CondBridge}/\textit{Frame} MTB-Avg 7.83/7.81).
Pedagogical-seed prompts help OOD specifically for \textit{ACE}, \textit{TextGrad}, and \textit{Frame}, while non-pedagogical seeds win on $R_\text{total}$ for \textit{ParetoGrad}/\textit{LeakShield}/\textit{GEPA}.
Education-specialization therefore yields gains in leak control and in selected OOD subscores, rather than a uniform advantage on $R_\text{total}$ (see \S\ref{sec:dissociation}).

\subsubsection*{RQ3: How do training-free methods compare to RL-trained approaches?}

Training-free optimization can match or exceed RL on aggregate reward: ParetoGrad reaches $R_\text{total}=0.719$ vs.\ RL Ped.\ Think R at 0.633, while using 100$\times$ less data (100 vs.\ 10K problems) and only inference-time compute on a single consumer GPU rather than the multi-GPU training stack required by RL.
RL still attains slightly tighter leak control under thinking (RL Ped.\ Think R leak 0.172 vs.\ ParetoGrad 0.252), and conversely training-free can exceed RL on $\Delta\text{Sol}$ (0.394 vs.\ 0.294).
The paradigms are complementary: RL provides tight behavioral control through weight updates, while training-free offers accessibility, interpretability, and immediate deployability with a much smaller compute footprint.

\begin{figure}[t]
\centering
\includegraphics[width=\columnwidth]{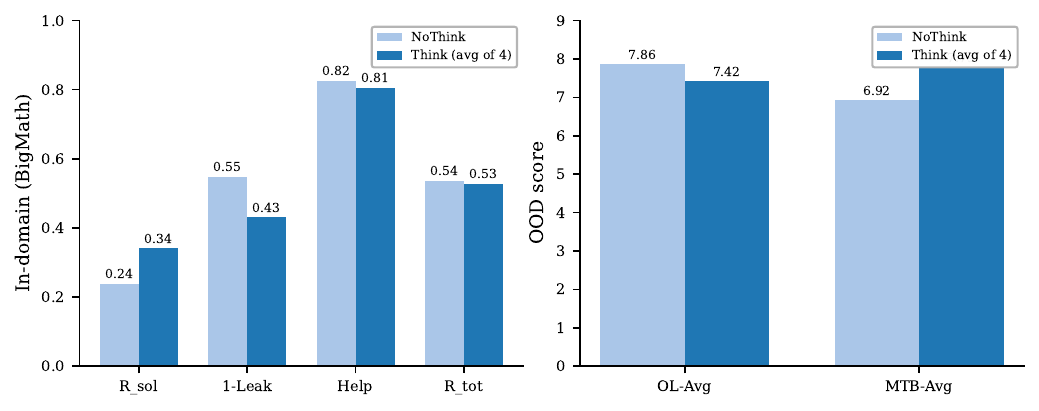}
\caption{\textit{Left} compares NoThink and Think (avg of 4 think conditions) on in-domain metrics (0-1 scale). Thinking modestly degrades leak control (1$-$Leak drops from 0.55 to 0.43). \textit{Right} shows OOD benchmark averages. Think improves MathTutorBench (+0.87) but hurts OpenLearnLM ($-$0.44), revealing a task-dependent reasoning mode effect.}
\label{fig:reasoning}
\end{figure}

Comparing Think and NoThink (Table~\ref{tab:main}, Figure~\ref{fig:reasoning} \textit{right}), thinking improves MathTutorBench but degrades OpenLearnLM.
Figure~\ref{fig:reasoning} \textit{left} shows that in-domain leak control degrades sharply under thinking.
This task-dependent reasoning mode effect is consistent across both training-free and RL paradigms \citep{lee2026pedagogicalrl-thinking}, suggesting it reflects a fundamental property of reasoning-enabled models rather than an artifact of the optimization method.
After the full sweep, the best condition per method is a Think variant for all 12 listed methods, with the choice between $R_\text{think}$ on/off and pedagogical/general seed differing across methods - encoded in Table~\ref{tab:main} by the indicator triplet rather than collapsed into a single representative condition.

\section{Analysis}

\begin{figure*}[t]
\centering
\includegraphics[width=\textwidth]{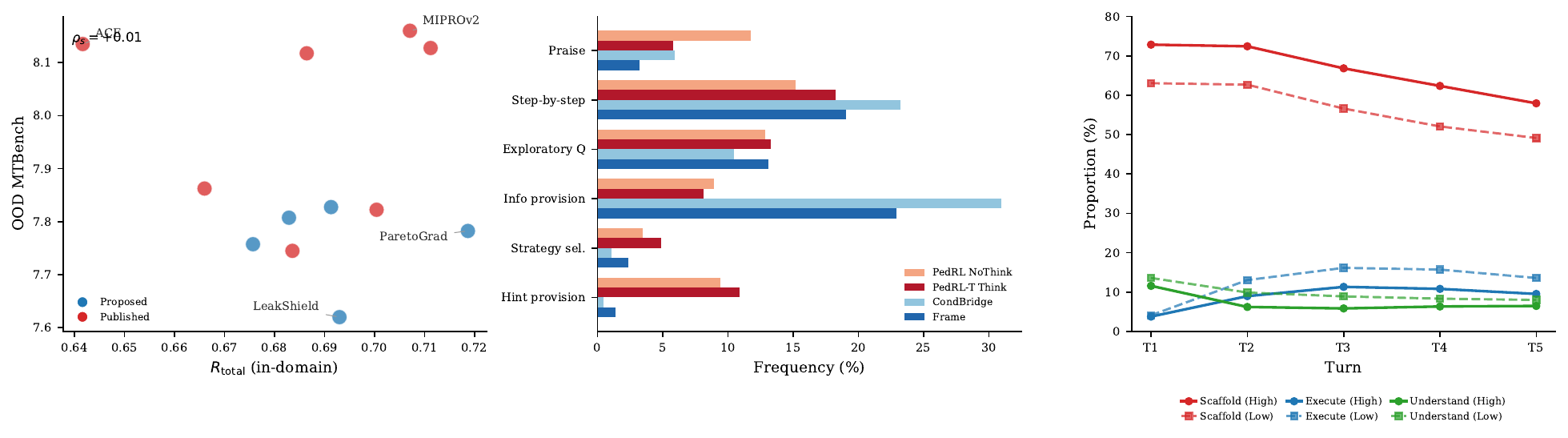}
\caption{\textit{Left} plots in-domain $R_\text{total}$ vs.\ OOD MTB-Avg across the 12 listed methods. Reward maximization is essentially uncorrelated with OOD MathTutorBench performance ($\rho\!=\!0.01$, $p\!=\!0.96$) and only weakly correlated with OpenLearnLM ($\rho\!=\!0.25$, $p\!=\!0.41$). \textit{Center} compares sentence-multilabel code frequency (\%) between RL-trained models \citep{dinucujianu2025pedagogicalrl,lee2026pedagogicalrl-thinking} and training-free methods (\textit{CondBridge}, \textit{Frame}). \textit{Right} reports Polya phase progression for top-6 vs.\ bottom-6 methods by $R_\text{total}$ (NoThink); high-performance methods sustain Scaffold codes throughout the dialog.}
\label{fig:analysis1}
\end{figure*}

\subsection{In-Domain vs.\ OOD Generalization}
\label{sec:dissociation}

Figure~\ref{fig:analysis1} reveals essentially no correlation between in-domain $R_\text{total}$ and OOD MathTutorBench-Avg across the 12 listed methods ($\rho = 0.01$, $p = 0.96$), and only a weak correlation with OpenLearnLM-Avg ($\rho = 0.25$, $p = 0.41$).
The most striking dissociations are: \textit{ACE} ranks 12th in-domain but 2nd on MTB (and 3rd on OL), \textit{LeakShield} ranks 5th in-domain but last (12th) on both MTB and OL, and \textit{ParetoGrad} ranks 1st in-domain but 9th on MTB and 7th on OL.
Codebook analysis suggests the mechanism: methods that minimize leak via tight scaffolding (ParetoGrad, LeakShield) optimize the in-domain reward rubric, while content-delivery-rich methods (ACE, MIPROv2, EvoPrompt) transfer their structured tutoring patterns more readily to held-out benchmarks.
This indicates that in-domain reward maximization and OOD generalization rely on fundamentally different pedagogical strategies.

\subsection{Trained vs.\ Training-Free Behavioral Comparison}
\label{sec:behav-comp}

We labeled 415{,}775 tutor sentences across 60 method-condition configurations (12 methods $\times$ 5 conditions) with GPT-4o-mini using the 82-code educational codebook from \citet{lee2026pedagogicalrl-thinking}.
Figure~\ref{fig:analysis1} \textit{center} compares two training-free methods (\textit{CondBridge}, \textit{Frame}) with RL baselines on six core codes (sentence-multilabel \%).
Training-free methods strongly suppress Praise (CondBridge 5.94\%, Frame 3.20\%; comparable to RL Ped.\ Think Reward 5.78\% and far below the unoptimized baseline 11.76\%), with \textit{MIPROv2} as a notable exception whose $R_\text{think}$-activated runs increase rather than reduce Praise (Appendix~\ref{app:miprov2-case}).
However, they redistribute behavior toward content delivery rather than explicit scaffolding: Information provision is 23-31\% under training-free versus 8-9\% under RL, while explicit Hint provision drops to 0.5-1.4\% versus 9-11\% under RL.
At the category level, this manifests as a 2-3$\times$ elevation of Mathematical Knowledge for Teaching (training-free 19-23\% vs.\ RL 7-8\%) and a ${\sim}10$pp deficit in Pedagogical Intent Utterance (training-free 46-55\% vs.\ RL 60-65\%), suggesting that prompt-level optimization recruits teaching-knowledge patterns where RL recruits intent-level scaffolding moves.

\subsection{Polya Phase Progression}
\label{sec:polya-phase}

We mapped the 82 codes to Polya's problem-solving phases and tracked progression across dialog turns (Figure~\ref{fig:analysis1} \textit{right}; top-6 vs.\ bottom-6 methods by $R_\text{total}$, NoThink).
Scaffold codes (hints, exploratory questions, prompts) dominate every turn (49-73\%), and high-performance methods maintain consistently higher Scaffold than low-performance methods (+9.8pp at T1, +8.8pp at T5).
Conversely, low-performance methods substitute direct Execution codes (calculation, step-by-step procedure) by 3-5pp across mid-late turns.
\emph{Sustained scaffolding}-continuing to guide rather than execute on the student's behalf-is the key behavioral differentiator between high- and low-performing optimization methods.

\begin{figure*}[t]
\centering
\includegraphics[width=\textwidth]{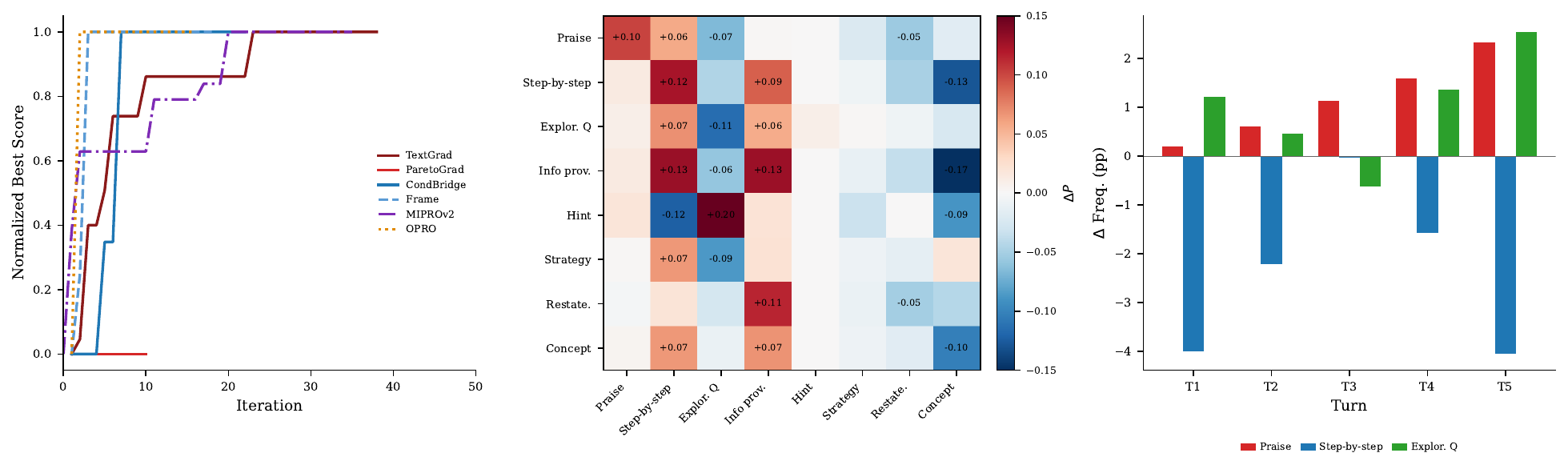}
\caption{\textit{Left} shows optimization convergence for 6 methods under a shared 500-evaluation budget; gradient methods converge within 10 iterations while dual-objective methods improve gradually. \textit{Center} displays code-to-code transition probability differences (high $-$ low). High-performance methods chain content delivery codes (Step-by-step, Information provision), while low-performance methods chain question codes (Exploratory question). \textit{Right} reports per-turn frequency differences between successful (top 25\%) and failed (bottom 25\%) dialogs in the high-performance group. Successful dialogs deploy fewer Step-by-step instructions and substitute more Exploratory questions and Praise toward the dialog's conclusion.}
\label{fig:analysis2}
\end{figure*}

\subsection{Optimization Convergence}

Figure~\ref{fig:analysis2} \textit{left} shows convergence patterns across methods.
All methods share a budget of 500 reward evaluations, but iterations differ because population methods (ParetoGrad, $N\!=\!5$) consume $\sim$50 evaluations per iteration while single-candidate methods (TextGrad) use $\sim$10.
Gradient-based methods converge rapidly within 10 iterations, while dual-objective methods (CondBridge, LeakShield) show slower but steadier improvement as they balance competing objectives.

\subsection{Transition Patterns}

Transition analysis (Figure~\ref{fig:analysis2} \textit{center}) reveals a code-type dependent pattern.
On \emph{question} codes, low-performance methods exhibit higher self-repetition: Exploratory question self-loop probability is 0.408 for low-performance vs.\ 0.357 for high-performance methods.
By contrast, on \emph{content-delivery} codes, high-performance methods show stronger self-chaining (Information provision 0.314 vs.\ 0.172; Step-by-step instruction 0.347 vs.\ 0.212), reflecting sustained explanation rather than fragmented question repetition.
The pedagogical signature of strong methods is thus a shift of self-repetition from question loops toward explanation chains.

\subsection{Success vs.\ Failure Dialogs}

Within the same high-performance methods, successful dialogs (top 25\% $R_\text{total}$) differ from failures across turns (Figure~\ref{fig:analysis2} \textit{right}).
Successful dialogs deploy fewer Step-by-step instructions early ($-$4.0pp at T1, $-$2.2pp at T2) and at closure ($-$4.0pp at T5), substituting more Exploratory questions (+1.2pp at T1, +2.6pp at T5) and more Praise (+2.3pp at T5).
The pattern suggests that successful tutoring elicits the student's understanding through questions rather than dictating procedure, and reserves praise for confirmed progress near the dialog's conclusion.
Decomposing this effect by reasoning mode and turn position (Appendix~\ref{app:praise-mode}) shows that the late-turn Praise contribution to $R_\text{total}$ is concentrated in NoThink dialogs at T4+, providing the behavioral substrate for the closing-turn pedagogical wrap-up.

\subsection{Cost Comparison}

PedagogicalRL-Thinking \citep{lee2026pedagogicalrl-thinking} requires substantial multi-GPU compute for hundreds of hours of GRPO training, while our approach uses only inference-time serving on a single consumer GPU during prompt optimization.
Training-free can match or exceed RL on aggregate reward (\textit{ParetoGrad} $R_\text{total}=0.719$ vs.\ RL Ped.\ Think R 0.633) with substantially lower GPU requirements, while producing interpretable, human-readable prompts that educators can directly inspect and modify.

\subsection{Qualitative Analysis}

While the preceding aggregate analyses characterize behavioral shifts in distribution, the excerpts here clarify what those shifts look like in dialog.
First, the leak-rate differences in Table~\ref{tab:main} and the Polya phase progression in \S\ref{sec:polya-phase} point to a shared mechanism: higher-performing methods avoid solution takeover while keeping the student responsible for the next mathematical move.
\textit{ParetoGrad} illustrates this delegation: ``Let's start by defining the number of red marbles as $R$. The problem gives us three relationships involving $R$. Can you write these relationships as equations?''
The tutor supplies variable naming and structural framing, but leaves equation formulation to the student \citep{wood1976tutoring}.
In pedagogical terms, this preserves the problem-solving cycle: the tutor supports entry into the task without collapsing Understand, Plan, and Execute into tutor-performed computation \citep{polya1945howto}.
The same pattern appears across both proposed and adapted methods, so the qualitative evidence is best read as a behavioral signature of successful prompt optimization rather than a proposed-versus-published contrast.

Second, the MKT/PIU redistribution discussed earlier appears qualitatively as a shift toward concept-rich explanation \citep{shulman1986knowledge,ball2008contentknowledge}.
\textit{CondBridge}, for example, frames inequalities and symmetry as a ``key insight'' for exploiting mathematical structure, explaining \emph{why} a representation is useful rather than merely \emph{which} operation to perform.
RL-trained tutors more often foreground interactional pacing through praise, questions, and hints.
Thus, training-free optimization and RL appear to emphasize different but compatible pedagogical resources: explanatory depth and scaffolding density.

Finally, some higher-performing dialogs elicit student-model turns with features of mathematical communication, including variable definition, function transformation, and explicit reasoning connectives.
After a structured \textit{ParetoGrad} exchange, the simulated student writes, ``let $a = \log_{10} 2$ \ldots\ Therefore $f(a) + f(-a) = \ldots$''
These are features of student-model generated discourse, not evidence of human learning.
Still, they suggest that future rewards could explicitly value transferable scaffolds, such as Polya-phase maintenance \citep{polya1945howto}, to better connect in-domain optimization with OOD teaching competencies.

\section{Discussion}

Training-free prompt optimization works because the prompt itself is an explicit pedagogical prior over the LLM's frozen distribution: short instructions can directly invoke teaching-knowledge patterns \citep{shulman1986knowledge,ball2008contentknowledge} that RL approximates only through many gradient updates over scalar rewards \citep{dinucujianu2025pedagogicalrl,lee2026pedagogicalrl-thinking}.
The observed MKT/PIU redistribution is consistent with this account: prompts recruit declarative resources, while RL shapes behavioral pacing \citep{chi2014icap}, making the two paradigms complements rather than substitutes.

The dissociation between in-domain reward and OOD generalization qualifies what ``best method'' means: tight-scaffolding methods win the in-domain rubric, while content-delivery-rich methods transfer better to held-out benchmarks \citep{macina2025mathtutorbench,lee2026openlearnlm}.
Since aggregate scores cluster tightly, \textit{ParetoGrad}'s value lies less in topping the table than in lacking weakness across the three reward components - Pareto balance is a more honest summary than scalar dominance for multi-objective evaluation.

For educators, training-free tutoring's most useful property is its artifact: a plain-text prompt that can be read, edited, and shared, rather than opaque weights.
Combined with the sustained scaffolding \citep{wood1976tutoring} and Polya-phase structure \citep{polya1945howto} visible in the dialogs, these prompts function more like inspectable instructional materials than fine-tuned models.

\section{Related Work}

\subsection{Prompt Optimization}
\label{sec:rel-po}

Prompt optimization refines black-box LLM prompts without weight updates via meta-prompting \citep{zhou2023ape,yang2024opro}, textual gradients \citep{pryzant2023protegi,yuksekgonul2024textgrad}, modular Bayesian programs \citep{khattab2024dspy,opsahlong2024miprov2}, evolutionary populations \citep{guo2024evoprompt}, or reflective evolution \citep{agrawal2026gepa,zhang2026ace}.
These methods have been benchmarked primarily on single-turn NLP tasks under a single scalar reward, with reflective evolution recently matching or exceeding reinforcement learning on agentic tasks.

\subsection{LLM-Based Math Tutoring}

LLM-based math tutors guide students through multi-turn dialog rather than supplying answers \citep{tack2022aiteacher,learnlmteam2025learnlm}, with scaffolding and answer-leak prevention as the defining design constraints \citep{chi2014icap}; recent work operationalizes this trade-off through dialog corpora \citep{macina2023mathdial}, multi-objective RL alignment \citep{dinucujianu2025pedagogicalrl,lee2026pedagogicalrl-thinking}, and OOD benchmarks \citep{macina2025mathtutorbench,lee2026openlearnlm}.
Our training-free alternative matches or exceeds the strongest RL-trained baseline using only inference-time prompts that educators can directly inspect, and enables systematic behavioral comparison over the same 82-code educational codebook \citep{lee2026pedagogicalrl-thinking}.

\section{Conclusion}

We investigated whether training-free prompt optimization can serve as a practical alternative to RL-based tutor alignment for math tutoring.
Adapting published prompt optimization methods to multi-turn educational dialog and proposing five education-specialized variants inspired by empirical failure analysis, we found that training-free methods match or exceed the strongest RL-trained baseline using only inference-time compute, while recruiting more teaching-knowledge patterns and less intent-level scaffolding than RL.
These results position training-free prompt optimization not as a cheaper substitute for RL but as a complementary paradigm that offers interpretability, accessibility, and the ability to explore diverse pedagogical strategies through method design alone.

\section{Limitations}

All tutoring dialogs use a simulated student (LLaMA-3.1-8B) rather than real learners; while standard in the field \citep{dinucujianu2025pedagogicalrl,lee2026pedagogicalrl-thinking}, student behavior may differ from authentic interactions.
Experiments are restricted to mathematics; generalization to other subjects is untested.
Evaluation relies on LLM-as-judge \citep{zheng2023llmjudge} without human evaluation.
We report single-run results without seed variance, though the breadth of comparison (792 runs) partially mitigates this.
Experiments use 7B/8B models; larger models may narrow the gap between methods.
Cross-paradigm comparison with RL uses different OOD benchmarks, limiting direct numerical comparison.
Codebook analysis uses GPT-4o-mini for automatic labeling (vs.\ GPT-5-mini in \citealt{lee2026pedagogicalrl-thinking}); a labeler-induced gap appears in absolute Praise rate, though the directional finding that training-free methods suppress Praise below the unoptimized baseline holds.

\section{Use of Generative AI}

We used Claude (Anthropic) to assist with drafting, editing, and code generation during the preparation of this manuscript.
All scientific claims, experimental design, and data analysis were conducted and verified by the authors.

\bibliography{custom}
\clearpage
\appendix

\section{Full Results}
\label{app:full-results}

Across the full experiment we designed 19 education-specialized prompt optimization methods targeting the three failure patterns identified in \S\ref{sec:main-results} (answer leakage, rigid instruction following, lack of pedagogical structure). The 5 representative methods (\textit{ParetoGrad}, \textit{CondBridge}, \textit{LeakShield}, \textit{Frame}, \textit{MetaBlend}) are described in \S\ref{sec:main-results} and listed in Table~\ref{tab:main}; the remaining 14 exploratory variants are described in \S\ref{app:methods}. All 19 methods were evaluated under the NoThink condition with the same multi-turn dialog simulation and multi-objective reward framework. Table~\ref{tab:full-indomain} presents in-domain BigMath results for the full set, sorted by $R_\text{total}$. $R_\text{total}$ uses the K=8 post-test $R_\text{sol}$ (consistent with Table~\ref{tab:main}); the $\Delta$Sol column is reported separately for reference. Two published methods (TextGrad, GEPA) are included with an asterisk for direct comparison under the same NoThink condition.

\begin{table}[h]
\centering
\scriptsize
\setlength{\tabcolsep}{2.5pt}
\caption{All 19 proposed methods on BigMath (NoThink).}
\label{tab:full-indomain}
\begin{tabular}{@{}lcccc@{}}
\toprule
\textbf{Method} & \textbf{$\Delta$Sol} & \textbf{Leak} & \textbf{Help} & $\boldsymbol{R_\text{total}}$ \\
\midrule
CondBridge    & .167 & \textbf{.204} & .848 & \textbf{.614} \\
TextGrad*     & .173 & .208 & .840 & .612 \\
ParetoGrad    & \textbf{.394} & .492 & \textbf{.849} & .594 \\
GEPA*         & .237 & .340 & .842 & .590 \\
HintGrad      & .263 & .370 & .846 & .589 \\
CurriculumOpt & .232 & .346 & .846 & .587 \\
DisCo         & .180 & .294 & .843 & .586 \\
HintChain     & .304 & .448 & .847 & .577 \\
PrincipleHint & .292 & .454 & .845 & .572 \\
DecompReward  & .167 & .320 & .840 & .571 \\
SokRat        & .284 & .446 & .845 & .570 \\
PopGrad       & .165 & .350 & .840 & .562 \\
Frame         & .195 & .392 & .840 & .558 \\
ThinkGuard    & .206 & .424 & .784 & .531 \\
LeakShield    & .167 & .508 & .782 & .489 \\
MetaBlend     & .117 & .492 & .801 & .486 \\
PromptDistill & .121 & .500 & .785 & .479 \\
Adversarial   & .162 & .700 & .821 & .437 \\
ContrastOpt   & .177 & .720 & .799 & .428 \\
AnchorBoost   & .105 & .668 & .794 & .421 \\
DualLoop      & .091 & .786 & .705 & .346 \\
\bottomrule
\multicolumn{5}{@{}l}{\scriptsize *Published method shown for reference.}
\end{tabular}
\end{table}

\section{Think Condition Comparison}
\label{app:think-conditions}

Table~\ref{tab:think-conditions} compares OOD performance across the four Think conditions for all 12 training-free methods.
Think Reward achieves the highest average (7.66), though differences across conditions are small (spread 0.08).
The best condition varies by method (marked with $\bigstar$), suggesting no single Think configuration is universally optimal.

\begin{table}[h]
\centering
\scriptsize
\setlength{\tabcolsep}{2.5pt}
\caption{OOD average across 4 Think conditions. $\bigstar$ = best condition per method. T = Think, PT = Ped.\ Think, NR = NoReward, R = Reward.}
\label{tab:think-conditions}
\begin{tabular}{@{}lcccc@{}}
\toprule
\textbf{Method} & \textbf{T-NR} & \textbf{T-R} & \textbf{PT-NR} & \textbf{PT-R} \\
\midrule
MIPROv2     & 7.76 & $\bigstar$7.92 & 7.90 & 7.91 \\
EvoPrompt   & 8.00 & $\bigstar$8.10 & 7.78 & 7.78 \\
ACE         & 7.98 & 7.82 & 7.87 & $\bigstar$8.00 \\
GEPA        & 7.54 & $\bigstar$7.76 & 7.59 & 7.61 \\
ParetoGrad  & 7.52 & $\bigstar$7.62 & 7.49 & 7.37 \\
OPRO        & 7.46 & $\bigstar$7.60 & 7.44 & 7.45 \\
TF-GRPO     & 7.54 & $\bigstar$7.60 & 7.51 & 7.54 \\
CondBridge  & 7.54 & 7.55 & $\bigstar$7.59 & 7.36 \\
MetaBlend   & 7.44 & 7.52 & $\bigstar$7.58 & 7.37 \\
Frame       & 7.44 & $\bigstar$7.55 & 7.49 & 7.53 \\
TextGrad    & 7.34 & 7.48 & $\bigstar$7.53 & 7.50 \\
LeakShield  & 7.37 & 7.39 & 7.47 & $\bigstar$7.52 \\
\midrule
\textit{Average} & \textit{7.58} & \textit{7.66} & \textit{7.60} & \textit{7.58} \\
\bottomrule
\end{tabular}
\end{table}

\section{Behavioral Analysis Details}
\label{app:behavior}

This section provides condition-level aggregates and additional cuts of the behavioral analysis in \S\ref{sec:behav-comp} and \S\ref{sec:polya-phase}. All training-free rows are 12-method averages under each condition; RL reference rows are from \citet{lee2026pedagogicalrl-thinking}.

\subsection{Labeling Methodology Details}
\label{app:labeling}

We use the 82-code educational codebook from \citet{lee2026pedagogicalrl-thinking}, organized into seven top-level categories: Mathematical Problem Solving (MPS), Mathematical Knowledge for Teaching (MKT), Cognition, Metacognition, Pedagogical Intent Utterance (PIU), Student Intent Utterance (SIU), and Affect/Discourse (A/D). Tutor responses are labeled with OpenAI GPT-4o-mini (\texttt{openai/gpt-4o-mini} via OpenRouter) at temperature 0. For thinking-mode trajectories, \texttt{<think>...</think>} blocks are stripped before sentence-level labeling, and sentences shorter than 15 characters are excluded. PedRL-Thinking originally reports labels obtained with GPT-5-mini; absolute rates therefore should not be directly compared across the two studies, but the within-study rank-based comparisons (Spearman) used in Appendix~\ref{app:praise-mode} are robust to labeler-induced offsets.

Two rate definitions appear in this work. The \emph{single-label rate} (used by aggregate category percentages following \citealt{lee2026pedagogicalrl-thinking}) is the fraction of sentences whose label set is exactly $\{c\}$; the \emph{multi-label rate} (used for dialog-level aggregation and for the core-code tables in this appendix) is the fraction of sentences whose label set contains $c$. Spearman correlation tests in Appendix~\ref{app:praise-mode} apply Benjamini-Hochberg correction across the $5\text{ conditions}\times 4\text{ reward components}$ grid (20 tests, $q<0.05$).

\begin{table}[h]
\centering
\scriptsize
\setlength{\tabcolsep}{4pt}
\caption{Labeled corpus statistics.}
\label{tab:labeling-stats}
\begin{tabular}{@{}lr@{}}
\toprule
\textbf{Quantity} & \textbf{Count} \\
\midrule
Labeled tutor sentences (60 method-condition cells) & 415{,}775 \\
Labeled dialogs with $R_\text{total}$ & 22{,}126 \\
Schoenfeld paragraphs (thinking only) & 30{,}954 \\
\bottomrule
\end{tabular}
\end{table}

\subsection{Category Distributions}

Table~\ref{tab:behav-cat-4} reports the 4-category code-instance \% per condition (the basis for the body's MKT/PIU finding), and Table~\ref{tab:behav-cat-7} expands the Interaction category into Cognition, Metacognition, Student Intent Utterance, and Affect/Discourse. The residual ${\sim}5{-}8$pp is dominated by Affect/Discourse, with small Cognition and Metacognition contributions that grow modestly under Think conditions.

\begin{table}[h]
\centering
\scriptsize
\setlength{\tabcolsep}{3pt}
\caption{Major category distribution (code-instance \%, 12-method average per condition). Rows sum to 100. NT = NoThink, T-NR = Think NoReward, T-R = Think Reward, PT-NR = Ped.\ Think NR, PT-R = Ped.\ Think R.}
\label{tab:behav-cat-4}
\begin{tabular}{@{}lcccc@{}}
\toprule
\textbf{Condition} & \textbf{MPS} & \textbf{MKT} & \textbf{PIU} & \textbf{Interaction} \\
\midrule
NT     & 20.29 & 19.31 & \textbf{54.81} &  5.59 \\
T-NR   & 20.37 & 22.57 & 49.37 &  7.68 \\
T-R    & 19.68 & 23.12 & 49.69 &  7.51 \\
PT-NR  & 24.03 & 22.57 & 46.22 &  7.17 \\
PT-R   & 23.73 & 22.77 & 47.27 &  6.23 \\
\midrule
\multicolumn{5}{@{}l}{\textit{RL reference \citep{lee2026pedagogicalrl-thinking}}} \\
RL NoThink     & 16.73 &  7.95 & 65.14 &  9.11 \\
RL Ped.\ Think & 25.06 &  7.15 & 59.54 &  6.97 \\
\bottomrule
\multicolumn{5}{@{}l}{\scriptsize MPS = Math Problem Solving, MKT = Math Knowledge for Teaching,} \\
\multicolumn{5}{@{}l}{\scriptsize PIU = Pedagogical Intent Utterance, Interaction = remaining categories.} \\
\end{tabular}
\end{table}

\begin{table}[h]
\centering
\scriptsize
\setlength{\tabcolsep}{2.5pt}
\caption{Extended 7-category distribution (code-instance \%, 12-method average per condition). Expands Table~\ref{tab:behav-cat-4} by splitting Interaction into Cognition, Metacognition, Student Intent Utterance, and Affect/Discourse.}
\label{tab:behav-cat-7}
\begin{tabular}{@{}lccccccc@{}}
\toprule
\textbf{Condition} & \textbf{MPS} & \textbf{MKT} & \textbf{Cog} & \textbf{Met} & \textbf{PIU} & \textbf{SIU} & \textbf{A/D} \\
\midrule
NT     & 20.29 & 19.31 & 0.69 & 0.80 & 54.81 & 0.27 & 3.83 \\
T-NR   & 20.37 & 22.57 & 1.91 & 1.82 & 49.37 & 0.31 & 3.65 \\
T-R    & 19.68 & 23.12 & 2.06 & 1.80 & 49.69 & 0.35 & 3.31 \\
PT-NR  & 24.03 & 22.57 & 1.59 & 1.91 & 46.22 & 0.36 & 3.31 \\
PT-R   & 23.73 & 22.77 & 1.51 & 1.82 & 47.27 & 0.20 & 2.69 \\
\bottomrule
\multicolumn{8}{@{}l}{\scriptsize Cog = Cognition, Met = Metacognition,} \\
\multicolumn{8}{@{}l}{\scriptsize SIU = Student Intent Utterance, A/D = Affect/Discourse.} \\
\end{tabular}
\end{table}

\subsection{Code Frequencies}

Table~\ref{tab:behav-codes} compares six core codes from the body's discussion against RL references, and Table~\ref{tab:behav-top15} expands the view to the 15 most frequent codes (by NoThink prevalence). Information provision, Step-by-step instruction, and Concept explanation dominate under NoThink (37.1\%, 30.0\%, 21.1\% of sentences), while Exploratory question is the most stable across conditions (17.5-18.7\% under Think variants).

\begin{table}[h]
\centering
\scriptsize
\setlength{\tabcolsep}{2.5pt}
\caption{Core code frequencies (sentence-multilabel \%, 12-method average per condition). Each entry is the fraction of sentences containing at least one instance of the code.}
\label{tab:behav-codes}
\begin{tabular}{@{}lcccccc@{}}
\toprule
\textbf{Condition} & \textbf{Praise} & \textbf{Step} & \textbf{ExplQ} & \textbf{Info} & \textbf{Hint} & \textbf{Strat} \\
\midrule
NT     & 5.48 & 29.95 & 18.73 & 37.10 & 1.30 & 0.04 \\
T-NR   & 5.99 & 14.47 & 18.31 & 22.28 & 1.45 & 0.10 \\
T-R    & 5.89 & 14.03 & 17.46 & 22.22 & 1.44 & 0.11 \\
PT-NR  & 4.02 & 16.66 & 13.57 & 23.20 & 1.32 & 0.07 \\
PT-R   & 3.35 & 18.61 & 14.04 & 27.28 & 1.11 & 0.08 \\
\midrule
\multicolumn{7}{@{}l}{\textit{RL reference \citep{lee2026pedagogicalrl-thinking}}} \\
RL NoThink     & 11.76 & 15.18 & 12.85 &  8.91 &  9.42 & 3.45 \\
RL Ped.\ Think &  5.78 & 18.22 & 13.27 &  8.12 & 10.89 & 4.89 \\
\bottomrule
\multicolumn{7}{@{}l}{\scriptsize Step = Step-by-step instruction, ExplQ = Exploratory question,} \\
\multicolumn{7}{@{}l}{\scriptsize Info = Information provision, Hint = Hint provision, Strat = Strategy selection.} \\
\end{tabular}
\end{table}

\begin{table}[h]
\centering
\scriptsize
\setlength{\tabcolsep}{3pt}
\caption{Top-15 codes by overall sentence-multilabel frequency, 12-method average per condition. Rows are sorted by NoThink frequency.}
\label{tab:behav-top15}
\begin{tabular}{@{}lccccc@{}}
\toprule
\textbf{Code} & \textbf{NT} & \textbf{T-NR} & \textbf{T-R} & \textbf{PT-NR} & \textbf{PT-R} \\
\midrule
Information provision           & 37.10 & 22.28 & 22.22 & 23.20 & 27.28 \\
Step-by-step instruction        & 29.95 & 14.47 & 14.03 & 16.66 & 18.61 \\
Concept explanation             & 21.10 & 17.09 & 16.80 & 16.44 & 17.74 \\
Exploratory question            & 18.73 & 18.31 & 17.46 & 13.57 & 14.04 \\
Application of rules/formulas   & 11.61 &  8.08 &  7.15 &  7.42 &  6.94 \\
Providing justification         & 11.45 & 13.28 & 12.35 & 12.25 & 11.36 \\
Performing calculations         &  6.93 &  4.24 &  3.84 &  4.07 &  3.75 \\
Open-ended question             &  6.36 &  4.60 &  3.74 &  3.83 &  3.78 \\
Praise                          &  5.48 &  5.99 &  5.89 &  4.02 &  3.35 \\
Encouraging participation       &  4.70 &  3.79 &  3.38 &  2.39 &  1.89 \\
Request for clarification       &  3.72 &  3.77 &  4.54 &  2.99 &  2.73 \\
Context check/Solution comp.    &  3.19 &  4.02 &  3.52 &  3.50 &  3.51 \\
Clarification                   &  3.04 &  5.26 &  4.58 &  5.84 &  5.74 \\
Emotional Support               &  2.94 &  2.39 &  2.37 &  1.61 &  1.13 \\
Use of examples                 &  2.71 &  2.69 &  2.64 &  2.67 &  2.12 \\
\bottomrule
\end{tabular}
\end{table}

\subsection{Polya Phase Distribution}

Table~\ref{tab:behav-polya-agg} summarises the Polya phase distribution over the full dialog, complementing the per-turn view in Figure~\ref{fig:analysis1} \textit{right}. Scaffold dominates every condition (58-63\%), while the pedagogical seed shifts mass from Interaction toward Understand (NT 8.2\% $\rightarrow$ PT-R 13.3\%).

\begin{table}[h]
\centering
\scriptsize
\setlength{\tabcolsep}{4pt}
\caption{Polya phase distribution over the full dialog (code-instance \%, 12-method average per condition).}
\label{tab:behav-polya-agg}
\begin{tabular}{@{}lccccc@{}}
\toprule
\textbf{Condition} & \textbf{Understand} & \textbf{Execute} & \textbf{Scaffold} & \textbf{Review} & \textbf{Interaction} \\
\midrule
NT     &  8.20 & 9.75 & 63.39 & 3.82 & 14.84 \\
T-NR   &  9.91 & 8.22 & 58.48 & 5.97 & 17.42 \\
T-R    & 10.01 & 7.71 & 59.03 & 5.81 & 17.44 \\
PT-NR  & 13.22 & 8.32 & 58.57 & 5.99 & 13.89 \\
PT-R   & 13.29 & 7.77 & 61.64 & 6.01 & 11.29 \\
\bottomrule
\end{tabular}
\end{table}

\subsection{Turn-Level Behavioral Diversity}

Table~\ref{tab:behav-turn} reports two diversity metrics per tutor turn: Shannon entropy of the code distribution and the mean number of distinct codes assigned. Think conditions raise entropy slightly above NoThink (1.51 nats $\rightarrow$ 1.59-1.62), consistent with the broader code mass shift visible in Tables~\ref{tab:behav-codes} and~\ref{tab:behav-top15}.

\begin{table}[h]
\centering
\scriptsize
\setlength{\tabcolsep}{4pt}
\caption{Turn-level behavioral diversity (12-method average per condition). Entropy is the Shannon entropy (nats) of the code distribution within a tutor turn; unique codes per turn is the mean number of distinct codes assigned to a tutor turn.}
\label{tab:behav-turn}
\begin{tabular}{@{}lcc@{}}
\toprule
\textbf{Condition} & \textbf{Entropy (nats)} & \textbf{Unique codes / turn} \\
\midrule
NT     & 1.505 & 7.82 \\
T-NR   & 1.621 & 8.11 \\
T-R    & 1.586 & 7.45 \\
PT-NR  & 1.609 & 8.21 \\
PT-R   & 1.570 & 7.72 \\
\bottomrule
\end{tabular}
\end{table}

\subsection{Schoenfeld Sanity Check}

Table~\ref{tab:behav-schoenfeld} reports the Schoenfeld 3-phase distribution over \texttt{<think>} content. This is included only as a sanity check against \citet{lee2026pedagogicalrl-thinking} Table 10; the body's behavioral analysis does not rely on it. Our Explore rate of 20-29\% diverges substantially from the 0.75\% reported under RL Ped.\ Think Reward, likely because our \texttt{<think>} blocks are short procedural traces rather than extended exploratory reasoning.

\begin{table}[h]
\centering
\scriptsize
\setlength{\tabcolsep}{4pt}
\caption{Schoenfeld 3-phase distribution (\%) of \texttt{<think>} content (paragraph-level, 12-method average per thinking condition).}
\label{tab:behav-schoenfeld}
\begin{tabular}{@{}lccc@{}}
\toprule
\textbf{Condition} & \textbf{Explore} & \textbf{General} & \textbf{Verify} \\
\midrule
T-NR   & 26.73 & 25.20 & 48.08 \\
T-R    & 28.52 & 24.20 & 47.28 \\
PT-NR  & 20.60 & 26.52 & 52.87 \\
PT-R   & 20.82 & 25.60 & 53.58 \\
\midrule
\multicolumn{4}{@{}l}{\textit{Reference \citep{lee2026pedagogicalrl-thinking}}} \\
RL Ped.\ Think R & 0.75 & 81.02 & 18.23 \\
\bottomrule
\end{tabular}
\end{table}

\subsection{Response Length and Math Content}

Table~\ref{tab:behav-wcmath} reports visible response length, thinking trace length, and the LaTeX math character fraction. NoThink responses are roughly 25\% shorter than Think visible responses (110.1 vs.\ ${\sim}145$ words) but contain a higher math character fraction (18.0\% vs.\ ${\sim}11\%$), reflecting the tendency of NoThink to deliver direct computational content.

\begin{table}[h]
\centering
\scriptsize
\setlength{\tabcolsep}{4pt}
\caption{Response length (mean words per turn) and math content (LaTeX math character fraction, \%) by condition (12-method average).}
\label{tab:behav-wcmath}
\begin{tabular}{@{}lccccc@{}}
\toprule
\textbf{Condition} & \textbf{Vis WC} & \textbf{Thk WC} & \textbf{Total WC} & \textbf{Vis Math\%} & \textbf{Thk Math\%} \\
\midrule
NT     & 110.1 &   0.0 & 110.1 & 17.97 & - \\
T-NR   & 146.9 & 201.3 & 348.2 & 11.17 & 0.93 \\
T-R    & 141.1 & 200.4 & 341.5 & 10.38 & 0.95 \\
PT-NR  & 150.2 & 204.0 & 354.3 & 10.30 & 1.02 \\
PT-R   & 145.9 & 201.7 & 347.6 & 11.27 & 1.02 \\
\bottomrule
\multicolumn{6}{@{}l}{\scriptsize Vis WC = visible response words, Thk WC = thinking-trace words.} \\
\end{tabular}
\end{table}

\subsection{Method Behavioral Clustering}

\begin{figure}[h]
\centering
\includegraphics[width=0.95\columnwidth]{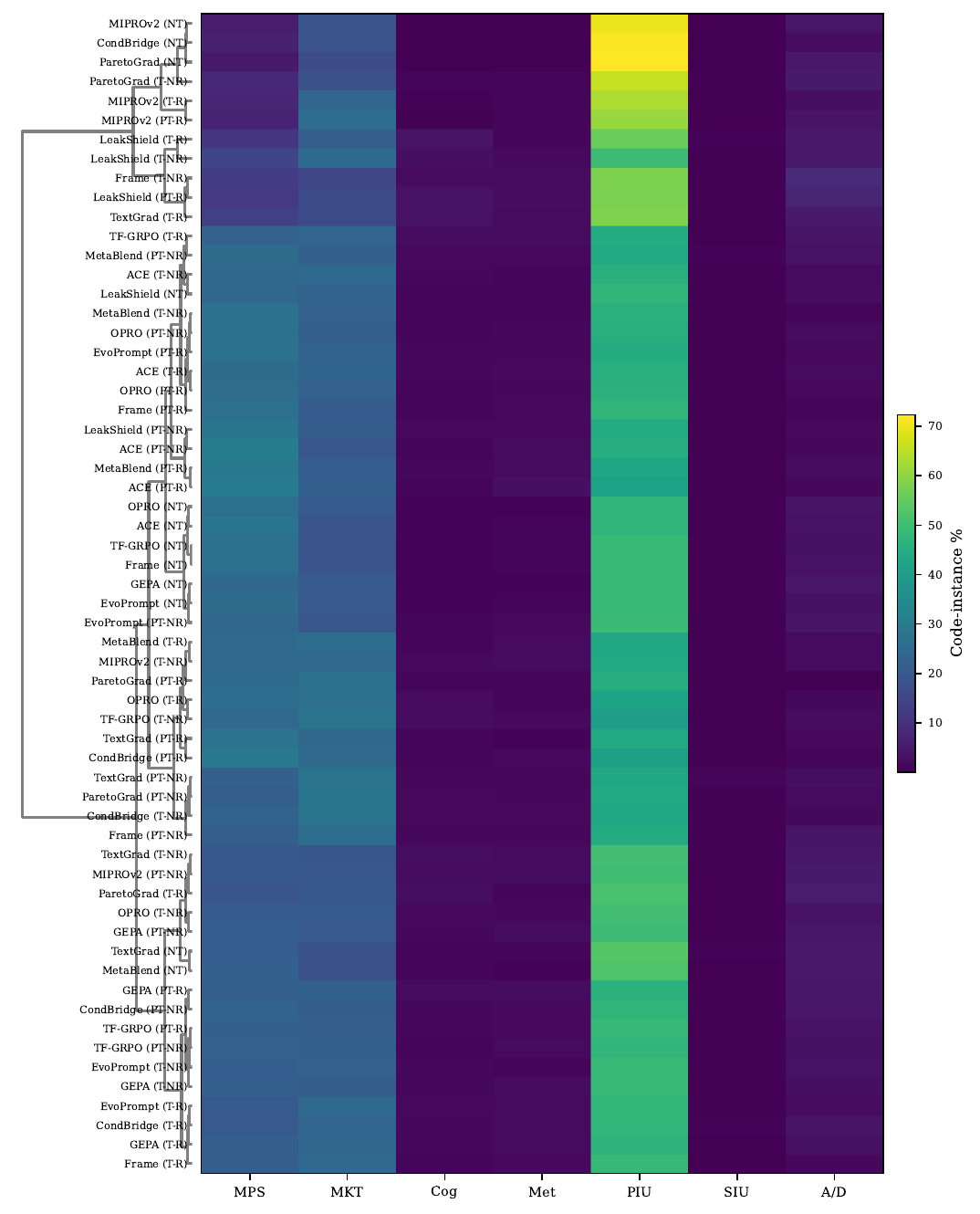}
\caption{Hierarchical clustering (Ward linkage on Euclidean distance) of 60 method-condition cells by their 7-category behavioral signature.}
\label{fig:method-clustermap}
\end{figure}

Figure~\ref{fig:method-clustermap} applies hierarchical clustering to the 60 method-condition cells using their 7-category behavioral signature. Cells with similar code-instance distributions group together, surfacing whether method identity or condition (NoThink vs.\ Think vs.\ Pedagogical seed) drives the dominant behavioral pattern.

\subsection{Condition Effect Analyses}

\begin{figure}[h]
\centering
\includegraphics[width=0.9\columnwidth]{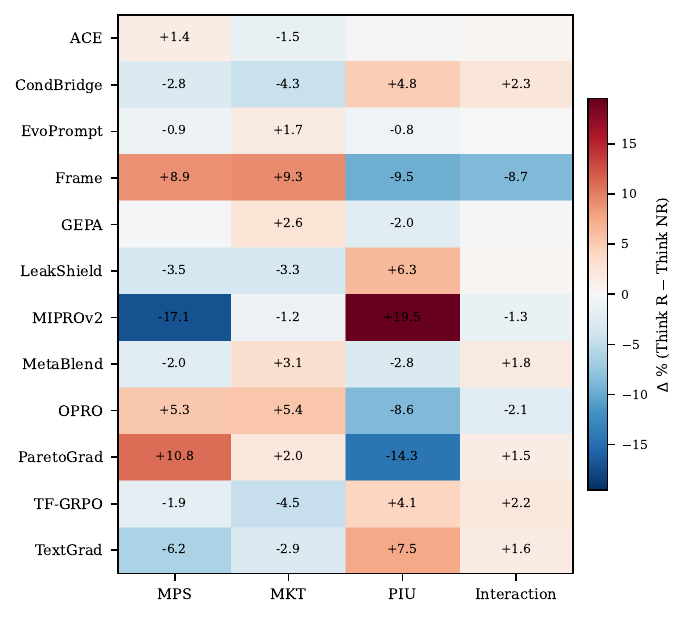}
\caption{Per-method effect of $R_\text{think}$ reward on the 4-category distribution (Think Reward $-$ Think NoReward, percentage points). Annotated cells have $|\Delta| \geq 0.5$pp.}
\label{fig:rthink-effect}
\end{figure}

\begin{figure}[h]
\centering
\includegraphics[width=0.9\columnwidth]{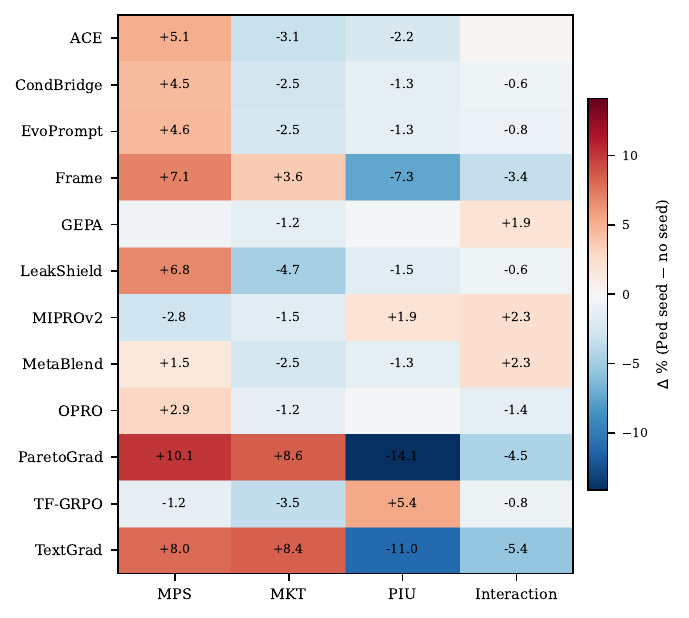}
\caption{Per-method effect of the pedagogical seed prompt on the 4-category distribution, averaged over the $R_\text{think}$ axis (Pedagogical seed $-$ no seed, percentage points). Annotated cells have $|\Delta| \geq 0.5$pp.}
\label{fig:pedseed-effect}
\end{figure}

Figures~\ref{fig:rthink-effect} and~\ref{fig:pedseed-effect} isolate the per-method effect of the two thinking-condition axes. The $R_\text{think}$ reward (Think Reward $-$ Think NoReward) produces small shifts ($|\Delta| \leq 2$pp in most cells), while the pedagogical seed produces larger and method-specific shifts, most prominently increased MPS and reduced PIU for several methods.

\subsection{Published vs.\ Proposed Group Comparison}

\begin{figure}[h]
\centering
\includegraphics[width=\columnwidth]{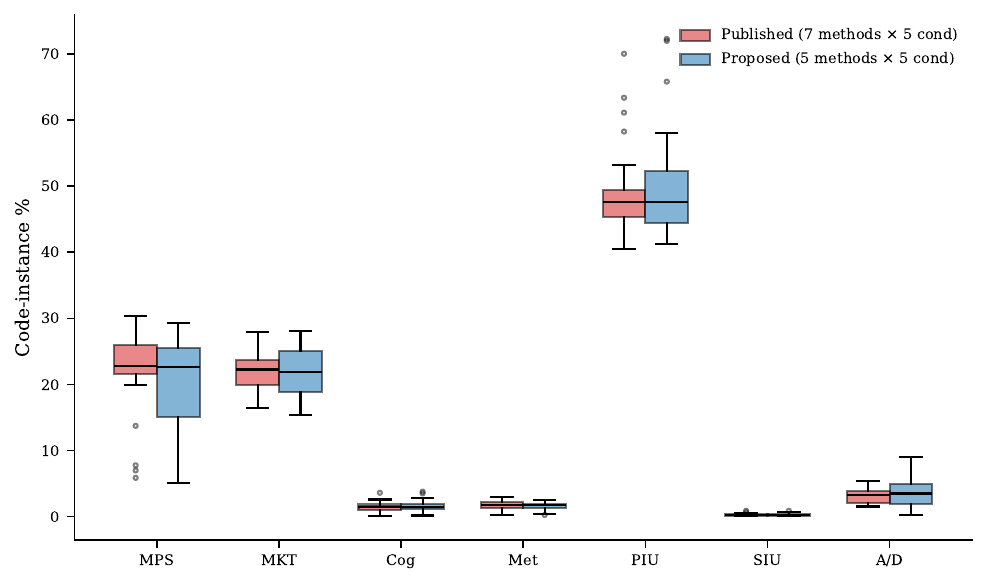}
\caption{Distribution of 7-category code-instance \% across method-condition cells, grouped by Published (7 methods $\times$ 5 conditions) vs.\ Proposed (5 methods $\times$ 5 conditions). Boxes show interquartile range with median; whiskers extend to 1.5$\times$IQR.}
\label{fig:pub-vs-prop}
\end{figure}

Figure~\ref{fig:pub-vs-prop} aggregates the 7-category distribution across all 5 conditions and compares the 7 adapted Published methods against our 5 Proposed methods. The two groups overlap substantially across all categories, indicating that the behavioral signature of training-free tutoring is largely driven by the prompt-optimization paradigm itself rather than by specific method design.

\subsection{Praise Mode-Specific Analysis}
\label{app:praise-mode}

Across $n=22{,}126$ labeled dialogs, we decompose the dialog-level Spearman correlation between Praise rate and reward components by reasoning condition and turn position. Table~\ref{tab:praise-mode-cond} shows that the positive correlation between Praise rate and $R_\text{total}$ is concentrated in the NoThink condition, while Think variants show null or slightly negative effects. Table~\ref{tab:praise-mode-turn} further localizes the effect to the dialog's closing turns (T4+), consistent with a late-turn motivational signal that carries measurable post-test contribution.

\begin{table}[h]
\centering
\scriptsize
\setlength{\tabcolsep}{3pt}
\caption{Per-condition Spearman correlation between dialog-level Praise rate and reward components. Significance after Benjamini-Hochberg correction over 20 tests. $^{*}$ $p<0.05$, $^{**}$ $p<0.01$, $^{***}$ $p<10^{-15}$.}
\label{tab:praise-mode-cond}
\begin{tabular}{@{}lccccc@{}}
\toprule
\textbf{Condition} & \textbf{n} & $\boldsymbol{\rho(R_\text{sol})}$ & $\boldsymbol{\rho(R_\text{leak})}$ & $\boldsymbol{\rho(R_\text{help})}$ & $\boldsymbol{\rho(R_\text{total})}$ \\
\midrule
NoThink  & 6{,}000 & $+0.126^{**}$ & $+0.016$       & $+0.077^{**}$  & $\mathbf{+0.106^{***}}$ \\
T-NR     & 4{,}131 & $+0.048$      & $-0.007$       & $+0.011$       & $+0.013$ \\
T-R      & 4{,}206 & $+0.056^{*}$  & $-0.010$       & $+0.023$       & $+0.026$ \\
PT-NR    & 3{,}900 & $-0.019$      & $+0.005$       & $+0.029$       & $+0.003$ \\
PT-R     & 3{,}889 & $-0.003$      & $-0.010$       & $+0.004$       & $-0.014$ \\
\bottomrule
\end{tabular}
\end{table}

\begin{table}[h]
\centering
\scriptsize
\setlength{\tabcolsep}{4pt}
\caption{Turn-position decomposition of Spearman correlation between dialog Praise rate and reward. The T4+ effect drives the bulk of the $R_\text{sol}$ contribution.}
\label{tab:praise-mode-turn}
\begin{tabular}{@{}lccc@{}}
\toprule
\textbf{Turn position} & $\boldsymbol{\rho(R_\text{sol})}$ & $\boldsymbol{\rho(R_\text{total})}$ & \textbf{Significance} \\
\midrule
T1-T3 (mid)  & $+0.046$ & $+0.062$ & $q<0.001$ \\
T4+ (close)  & $+0.093$ & $\mathbf{+0.110}$ & $q<10^{-28}$ \\
\bottomrule
\end{tabular}
\end{table}

\subsection{MIPROv2 Reward-Driven Behavioral Shift}
\label{app:miprov2-case}

The Praise-suppression generalization in \S\ref{sec:behav-comp} is a method-level aggregate. \textit{MIPROv2} is a notable exception: when $R_\text{think}$ is activated, its Praise rate increases by $+8.49$pp, its Pedagogical Intent Utterance share by $+19.5$pp, and its turn-level code-distribution entropy collapses by $-0.91$ bits - each the most extreme value among the 12 listed methods (\textit{ParetoGrad} moves in the opposite direction with $\Delta\text{Praise}=-5.03$pp). Table~\ref{tab:miprov2-outlier} lists four behavioral category cells where MIPROv2 exceeds the 12-method mean by $\geq 2\sigma$, all sharing a ``high PIU, low MPS'' signature consistent with a reward-driven tilt toward direct telling at the expense of Socratic questioning. These cells contribute to MIPROv2's competitive $R_\text{total}=0.707$ despite the behavioral divergence, illustrating that scalar reward alone can mask qualitatively different policies.

\begin{table}[h]
\centering
\scriptsize
\setlength{\tabcolsep}{4pt}
\caption{Behavioral category cells where \textit{MIPROv2} exceeds the 12-method mean by $\geq 2\sigma$. PIU = Pedagogical Intent Utterance, MPS = Mathematical Problem Solving.}
\label{tab:miprov2-outlier}
\begin{tabular}{@{}llcc@{}}
\toprule
\textbf{Condition} & \textbf{Category} & \textbf{MIPROv2 (\%)} & \textbf{12-method avg (\%)} \\
\midrule
Think Reward      & PIU & 63.4 & 49.7 \\
Think Reward      & MPS &  7.8 & 19.7 \\
Ped Think Reward  & PIU & 61.1 & 47.2 \\
Ped Think Reward  & MPS &  7.0 & 23.7 \\
\bottomrule
\end{tabular}
\end{table}

\subsection{Method-Level Seed Effect Breakdown}
\label{app:seed-effect}

Table~\ref{tab:seed-effect-method} expands the aggregate seed analysis to a per-method breakdown of the Polya pedagogical seed effect (pedagogical condition $-$ general condition, NoThink) on the two largest behavioral categories. Negative $\Delta$PIU combined with positive $\Delta$MPS indicates a shift away from explicit telling toward Socratic question-driven instruction. As a descriptive observation (not a statistically inferential claim given $n=12$), four of the five proposed methods (\textit{ParetoGrad}, \textit{Frame}, \textit{LeakShield}, \textit{MetaBlend}) show stronger seed-driven PIU/MPS reallocation than any of the seven published methods, consistent with the design intent that the proposed methods explicitly embed pedagogical priors (\S\ref{sec:proposed}).

\begin{table}[h]
\centering
\scriptsize
\setlength{\tabcolsep}{4pt}
\caption{Per-method Polya pedagogical seed effect (pedagogical $-$ general, NoThink condition, percentage points). Rows are sorted by $\Delta$PIU.}
\label{tab:seed-effect-method}
\begin{tabular}{@{}llrr@{}}
\toprule
\textbf{Method} & \textbf{Source} & $\boldsymbol{\Delta\text{PIU}}$ & $\boldsymbol{\Delta\text{MPS}}$ \\
\midrule
ParetoGrad & Proposed & $-22.0$ & $+13.9$ \\
Frame      & Proposed & $-13.6$ & $+9.2$ \\
TextGrad   & Published & $-7.6$ & $+2.1$ \\
OPRO       & Published & $-5.1$ & $+6.2$ \\
LeakShield & Proposed & $-4.8$  & $+13.2$ \\
MetaBlend  & Proposed & $-1.9$  & $-1.4$ \\
ACE        & Published & $-0.9$ & $+6.1$ \\
GEPA       & Published & $+0.3$ & $-0.8$ \\
EvoPrompt  & Published & $+0.7$ & $+2.4$ \\
CondBridge & Proposed & $+4.1$  & $-0.0$ \\
MIPROv2    & Published & $+6.1$ & $-4.9$ \\
TF-GRPO    & Published & $+7.1$ & $-2.0$ \\
\bottomrule
\end{tabular}
\end{table}

\subsection{Method-Level Reward Effect Breakdown}
\label{app:reward-effect}

Complementing the MIPROv2 case study in Appendix~\ref{app:miprov2-case}, Table~\ref{tab:reward-effect-method} reports the per-method $R_\text{think}$ reward effect (reward on $-$ reward off, Think condition) on three behavioral indicators: Praise rate (sentence-multilabel \%), PIU share (code-instance \%), and turn-level code-distribution entropy (bits). The reward signal acts differently across methods in both magnitude and direction: \textit{MIPROv2} sharpens to a single high-Praise high-PIU pattern (entropy collapse), while \textit{ParetoGrad} moves in the opposite direction, suppressing Praise and PIU while broadening the code distribution. Aggregate $R_\text{total}$ does not distinguish these qualitatively different policies.

\begin{table}[h]
\centering
\scriptsize
\setlength{\tabcolsep}{3pt}
\caption{Per-method $R_\text{think}$ reward effect (reward on $-$ reward off, Think condition). Rows sorted by $\Delta$Praise.}
\label{tab:reward-effect-method}
\begin{tabular}{@{}llrrr@{}}
\toprule
\textbf{Method} & \textbf{Source} & $\boldsymbol{\Delta\text{Praise}}$ (pp) & $\boldsymbol{\Delta\text{PIU}}$ (pp) & $\boldsymbol{\Delta\text{entropy}}$ (bits) \\
\midrule
MIPROv2    & Published & $+8.49$ & $+19.5$ & $-0.91$ \\
ACE        & Published & $+2.19$ & $-0.3$  & $-0.03$ \\
CondBridge & Proposed  & $+0.28$ & $+4.8$  & $-0.01$ \\
MetaBlend  & Proposed  & $+0.17$ & $-2.9$  & $+0.00$ \\
LeakShield & Proposed  & $+0.11$ & $+6.4$  & $-0.08$ \\
OPRO       & Published & $+0.03$ & $-8.5$  & $-0.66$ \\
EvoPrompt  & Published & $-0.61$ & $-0.8$  & $+0.18$ \\
GEPA       & Published & $-0.76$ & $-2.1$  & $-0.20$ \\
TF-GRPO    & Published & $-0.82$ & $+4.1$  & $+0.41$ \\
Frame      & Proposed  & $-1.33$ & $-9.5$  & $-0.22$ \\
TextGrad   & Published & $-3.89$ & $+7.5$  & $-0.22$ \\
ParetoGrad & Proposed  & $-5.03$ & $-14.4$ & $+0.64$ \\
\bottomrule
\end{tabular}
\end{table}

\subsection{CondBridge Prompt-Evolution Artifact}
\label{app:condbridge-artifact}

\textit{CondBridge} under NoThink shows a Praise rate of 10.33\%, the only cell among the 60 method-condition combinations to exceed 7\% (other NoThink cells fall in the 3-7\% range; cf.\ Table~\ref{tab:behav-codes}). Inspection of the optimization trajectory reveals that the seed prompt contains no explicit praise template - the word ``encourage'' appears in instruction context but no praise phrase is templated. During prompt-evolution mutation an explicit closing template is introduced:
\begin{quote}
``Conclude your response with an encouraging statement, such as `You're doing a great job!' or `Keep up the good work!'\,''
\end{quote}
This template appears in the optimized best prompt but is absent from the seed. The 10.33\% Praise rate is therefore a \emph{prompt-evolution artifact}: a localized template insertion that survived the dual-condition objective rather than a direct consequence of the objective itself. In contrast to the MIPROv2 case (Appendix~\ref{app:miprov2-case}), which illustrates reward-response hacking, this CondBridge case illustrates mutation-injection hacking, indicating that prompt-level optimization can introduce verbal-praise patterns through at least two distinct mechanisms.

\subsection{Thinking-Mode Entropy Effect}
\label{app:thinking-entropy}

Thinking-mode trajectories show higher behavioral entropy and a more uniform distribution across methods than NoThink trajectories (Table~\ref{tab:thinking-entropy}). The mean turn-level Shannon entropy (computed over the per-turn distribution of assigned codes) increases from 2.644 bits under NoThink to 2.977 bits under Think (4-condition average), and the method-level entropy standard deviation contracts from 0.406 to 0.246 bits, indicating that thinking partially normalizes behavioral diversity across methods. The reward-driven entropy shifts in Appendix~\ref{app:reward-effect} (\textit{ParetoGrad} $+0.64$ bits vs.\ \textit{MIPROv2} $-0.91$ bits) operate on top of this thinking-mode baseline.

\begin{table}[h]
\centering
\scriptsize
\setlength{\tabcolsep}{3pt}
\caption{Turn-level behavioral entropy (bits) under NoThink and Think conditions, 12-method average. \textit{Method std} = std of entropy across methods at fixed condition.}
\label{tab:thinking-entropy}
\begin{tabular}{@{}lrrr@{}}
\toprule
\textbf{Condition} & \textbf{Codes/turn} & \textbf{Entropy} & \textbf{Method std} \\
\midrule
NoThink      & 7.95  & 2.644 & 0.406 \\
Think (avg)  & 10.95 & 2.977 & 0.246 \\
\bottomrule
\end{tabular}
\end{table}

\section{Method Details}
\label{app:methods}

This section provides detailed descriptions of the 14 proposed methods not covered in the main text. All methods use the same multi-turn dialog simulation and multi-objective reward framework described in \S2.

\subsection{Pedagogical Scaffolding}

\textit{HintChain} ($R_\text{total}=0.577$) optimizes a progressive hint sequence where each hint builds on the previous one with increasing specificity. The optimization evaluates whether students progress after each hint level, adjusting the structure to minimize hints needed while maximizing solve rate. Optimization runs for 500 metric calls with minibatch size 10.

\textit{HintGrad} ($R_\text{total}=0.589$) applies TextGrad-style gradient feedback specifically to the hint-giving portions of the tutor prompt. Gradients target cases where hints were too vague (no student progress) or too revealing (answer leaked), refining the balance between guidance and discovery.

\textit{SokRat} ($R_\text{total}=0.570$) optimizes Socratic questioning strategies by evaluating question quality and student response patterns. The optimizer adjusts questioning patterns to maximize engagement and problem-solving progress, using dialog turn analysis as feedback.

\subsection{Solution-Leak Prevention}

\textit{Adversarial} ($R_\text{total}=0.437$) generates red-team prompt mutations designed to expose leakage vulnerabilities. Each iteration, an adversary LLM modifies the tutor prompt to induce answer leakage. Successfully leaked cases are used to patch the prompt, creating robustness against leakage failure modes.

\textit{ThinkGuard} ($R_\text{total}=0.531$) monitors chain-of-thought traces for premature solution disclosure. When the thinking trace contains the final answer before the tutor has guided the student, ThinkGuard adds thinking-specific anti-leak constraints to the prompt.

\subsection{Dual-Objective}

\textit{DualLoop} ($R_\text{total}=0.346$) combines an inner loop (TextGrad gradient descent at every iteration) with an outer loop (behavioral proxy analysis every 5 iterations). The outer loop extracts three proxy metrics from dialog transcripts: question ratio (fraction of tutor turns containing questions), average disclosure level (0--4 scale), and average turn count. When proxies deviate from target ranges (e.g., 10--20\% question ratio, disclosure $\leq 1.5$), priority signals steer optimization toward desirable behavioral patterns.

\textit{CurriculumOpt} ($R_\text{total}=0.587$) applies curriculum learning to prompt optimization: early iterations optimize on easier problems (higher baseline solve rate), with difficulty gradually increasing. This prevents premature convergence on strategies that only work for simple problems.

\subsection{Meta / Distillation}

\textit{PromptDistill} ($R_\text{total}=0.479$) analyzes best-performing prompts from multiple optimization methods, identifies common successful patterns, and distills them into a single compact prompt. A refinement phase then optimizes this distilled prompt using TextGrad for 300 iterations.

\textit{DecompReward} ($R_\text{total}=0.571$) decomposes the aggregate reward into fine-grained pedagogical sub-components (e.g., question quality, hint specificity, encouragement appropriateness) and optimizes each sub-component with targeted gradient feedback before re-composing the full prompt.

\subsection{Contrastive Analysis}

\textit{DisCo} ($R_\text{total}=0.586$) mines dialog transcripts for recurring patterns associated with high or low rewards, then uses these discovered patterns as optimization heuristics. Successful patterns are reinforced in the prompt while failure patterns are explicitly discouraged.

\textit{AnchorBoost} ($R_\text{total}=0.421$) identifies anchor prompt segments from early optimization that consistently contribute to high performance, freezes these segments, and applies gradient-based optimization only to the remaining prompt regions. This preserves verified pedagogical strategies while exploring improvements.

\textit{ContrastOpt} ($R_\text{total}=0.428$) evaluates each candidate prompt under both NoThink and Think conditions on the same examples each iteration and uses per-example gap analysis as the optimization signal. The combined score $0.5 R^\text{NT} + 0.5 R^\text{TH} - 0.3 |R^\text{NT} - R^\text{TH}|$ penalizes condition-dependent inconsistency, and the optimizer receives detailed feedback on examples where one condition succeeded while the other failed.

\subsection{Population-Gradient Hybrids}

\textit{PopGrad} ($R_\text{total}=0.562$) maintains a population of prompts and uses TextGrad gradient information to guide mutation operators. Unlike standard evolutionary approaches with random mutations, PopGrad directs mutations based on gradient feedback, combining exploration (population diversity) with exploitation (gradient direction).

\textit{PrincipleHint} ($R_\text{total}=0.572$) generates prompts from explicit pedagogical principles (e.g., Polya's problem-solving steps, zone of proximal development) and integrates progressive hint structures. The optimization adjusts the balance between principle-guided instruction and adaptive hint giving across 500 metric calls.

\end{document}